\documentclass[conference]{IEEEtran}
\IEEEoverridecommandlockouts
\usepackage{caption} 
\usepackage{multirow}
\usepackage{booktabs}
\usepackage{subfig}
\usepackage{cite}
\usepackage{amsmath,amssymb,amsfonts}
\usepackage{algorithmic}
\usepackage{graphicx}
\usepackage{booktabs}
\usepackage{amsmath}
\usepackage{threeparttable}
\usepackage{multirow}
\usepackage{graphicx}
\usepackage{textcomp}
\usepackage{xcolor}
\def\BibTeX{{\rm B\kern-.05em{\sc i\kern-.025em b}\kern-.08em
    T\kern-.1667em\lower.7ex\hbox{E}\kern-.125emX}}
\begin{document}

\title{LESS YET ROBUST: CRUCIAL REGION SELECTION FOR SCENE RECOGNITION}

\author{
\IEEEauthorblockN{Jianqi Zhang}
\IEEEauthorblockA{
\textit{Institute of Software, Chinese Academy of Sciences}\\
Beijing, China \\
jluzhangjianqi@163.com}

\IEEEauthorblockN{Jingyao Wang}
\IEEEauthorblockA{
\textit{Institute of Software, Chinese Academy of Sciences} \\
Beijing, China \\
jingyao\_wang0728@163.com}

\IEEEauthorblockN{Changwen Zheng}
\IEEEauthorblockA{
\textit{Institute of Software, Chinese Academy of Sciences} \\
Beijing, China \\
changwen@iscas.ac.cn}

\and

\IEEEauthorblockN{Mengxuan Wang}
\IEEEauthorblockA{
\textit{South China University of Technology} \\
Guangzhou, China \\
wimannix@mail.scut.edu.cn}

\IEEEauthorblockN{Lingyu Si}
\IEEEauthorblockA{
\textit{Institute of Software, Chinese Academy of Sciences} \\
Beijing, China \\
lingyu@iscas.ac.cn}

\IEEEauthorblockN{Fanjiang Xu}
\IEEEauthorblockA{
\textit{Institute of Software, Chinese Academy of Sciences} \\
Beijing, China \\
fanjiang@iscas.ac.cn}
}
\maketitle

\begin{abstract}
Scene recognition, particularly for aerial and underwater images, often suffers from various types of degradation, such as blurring or overexposure. Previous works that focus on convolutional neural networks have been shown to be able to extract panoramic semantic features and perform well on scene recognition tasks. However, low-quality images still impede model performance due to the inappropriate use of high-level semantic features. To address these challenges, we propose an adaptive selection mechanism to identify the most important and robust regions with high-level features. Thus, the model can perform learning via these regions to avoid interference. implement a learnable mask in the neural network, which can filter high-level features by assigning weights to different regions of the feature matrix. We also introduce a regularization term to further enhance the significance of key high-level feature regions. Different from previous methods, our learnable matrix pays extra attention to regions that are important to multiple categories but may cause misclassification and sets constraints to reduce the influence of such regions.This is a plug-and-play architecture that can be easily extended to other methods. Additionally, we construct an Underwater Geological Scene Classification dataset to assess the effectiveness of our model. Extensive experimental results demonstrate the superiority and robustness of our proposed method over state-of-the-art techniques on two datasets.
\end{abstract}


\section{Introduction}
Scene recognition has attracted extensive attention from the scientific community and the public, which aims to classify a brief interpretation of the scene in the image. It has extensive application in environmental monitoring \cite{Zhao2018}, rescue \cite{2022Bh}, marine archaeology \cite{2021GU} and serves significant value in military and aviation activities \cite{2022Qi}. 
Unfortunately, recent scene recognition tasks mainly focus on land scene classification using high-resolution color remote images, such as \cite{aid,Eurosat,UCM}, which may still face challenges in real-world applications, e.g., underwater \cite{Liu2016,Yuh2011,underwater2022utd} and remote sensing \cite{Chaib2017,wang2024towards} scenarios. 

\begin{figure}
    \centering
    \includegraphics[width=1\linewidth]{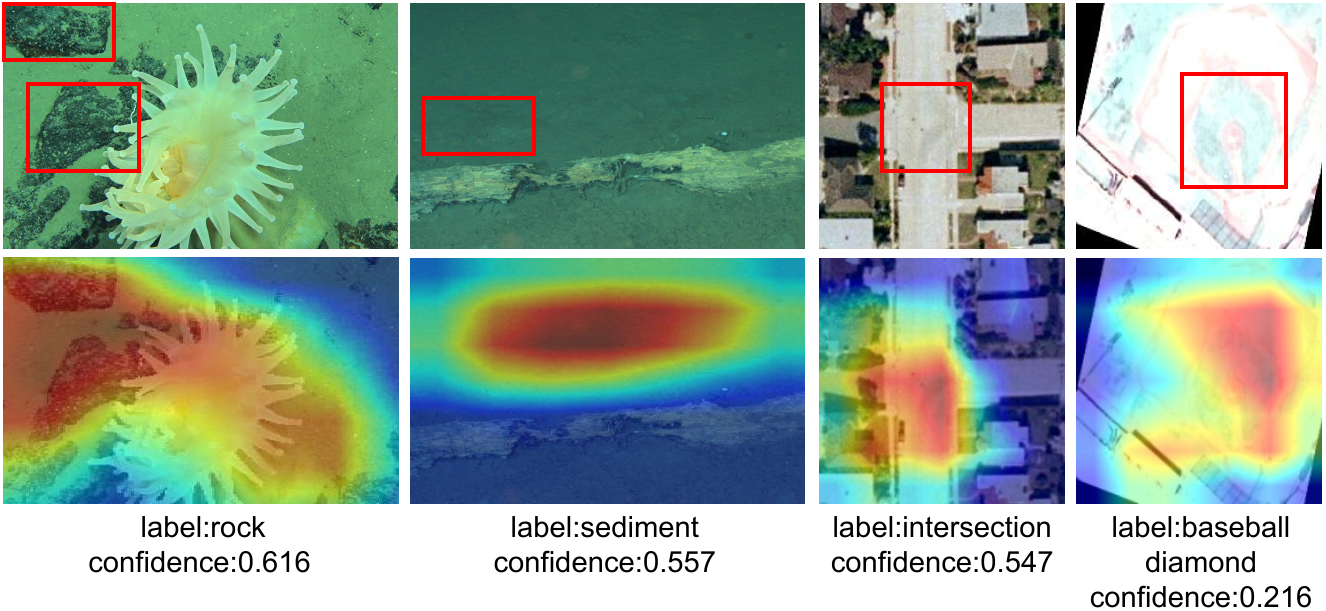}
    \caption{The first row shows the input images, where the red boxes indicate the areas sufficient to determine the class of the image. The second row (obtained through Grad-CAM \cite{Grad2017}) shows the areas that ResNet18 focuses on when predicting these images. The labels of the images and the confidence score for these labels are shown at the bottom of the images. It is evident that the attention areas of ResNet18 are significantly larger than the regions within the red boxes.}
    \label{fig:visualize}
\end{figure} 

Recently, deep neural network models based on CNNs have achieved great success in visual classification tasks and have begun to be introduced to scene recognition tasks \cite{2019H,2019ww,2015cheng,wang2022ssd}. With the advancement of technology, models with deeper CNN architectures are generally implemented in the scene recognition field to extract invariant semantic features for classification. For instance, ResNet \cite{2016Deep} and its variants, pre-trained on ImageNet \cite{2009L}, have shown great potential in downstream visual tasks. They primarily serve as feature extractors to obtain semantic features, which are then used as input for the classification head \cite{Liu2024SCECNetSF,s2023,2017MMMM,Phapale2021}.

\begin{figure*}[htbp]
    \centering
    \includegraphics[width=1\linewidth]{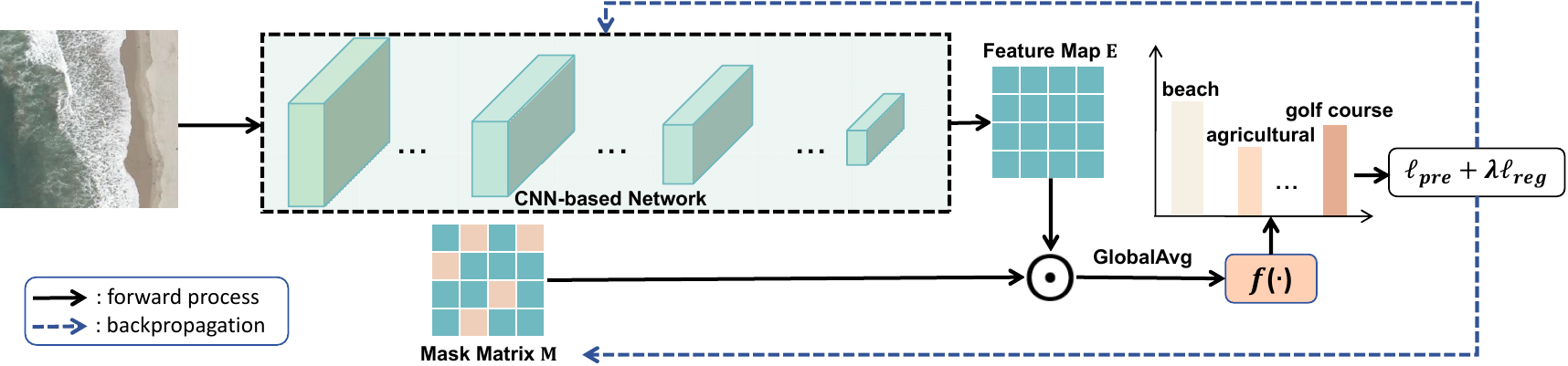}
    \caption{The proposed method's overall framework. Typically, features near the input end are considered to be low-level semantic information, containing fine-grained semantics, while features near the output end are considered to be high-level semantic information, containing coarser semantics.}
    \label{scheme}
\end{figure*}

However, in real-world applications, scene recognition tasks can be affected by factors such as object occlusion, lighting blur, and overexposure. Previous CNNs-based methods, which learn from all regions of the image, may mistakenly incorporate these distracting elements into the decision-making process. Taking the CNNs-based model, ResNet-18, as an example, we visualize the attention regions that the model focuses on when predicting the labels of four images in Fig. \ref{fig:visualize}. The red box area in Fig. \ref{fig:visualize} is sufficient to determine the image category, but ResNet18 focuses on regions significantly beyond this area. For example, in the first column, the model unnecessarily focuses on the sea anemone obscuring the scene, and in the fourth column, it concentrates excessively on areas unrelated to the label. This focus on irrelevant areas introduces additional noise, leading to performance degradation. Therefore, we propose focusing on more discriminative and robust regions of the image, allowing the model to learn from these areas to eliminate the impact of noise or occlusion on decision-making in practical applications. It is important to note that focusing solely on regions crucial for decision-making is not enough, as some regions that are important for multiple categories may lead to errors when the model tries to distinguish between them (as shown in the third column of Fig. \ref{fig:visualize}). Fig. \ref{fig:visualize_all} provides further analysis of Fig. \ref{fig:visualize}.

Based on this insight, we develop a novel method that propels the model to focus on less but robust high-level feature regions. The model learned via these crucial semantic features will achieve higher performance. Specifically, we adopt a CNN-based extractor to obtain semantic features and then use a learnable mask matrix to locate important features. Consequently, this learnable matrix first re-assigns weights to different regions to strengthen the model's focus on decision-related semantics, and then introduces mask constraints to eliminate the semantic influence of category sharing, allowing the model to make decisions based on fewer but more robust regions. Our main contributions are in four aspects:
\begin{itemize}
\item We provide a straightforward and valid feature selection strategy to propel the model to catch less yet crucial high-level semantic feature regions.
\item We propose an importance regularization term that encourages sparsity in the mask by pushing more elements to zero, thus reducing the influence of unrelated high-level semantic feature regions.
\item A new dataset, the Underwater Geological Scene (UGS), is developed for classifying high-resolution submarine sensing images, alleviating the scarcity of underwater scene recognition tasks. The dataset aims to classify underwater geological categories, including two common submarine scenes.
\item This proposed method can be easily extended to CNN-based models, and extensive experimental results validate its superiority and effectiveness. 
\end{itemize}

\section{Methodology}

In this section, we introduce the proposed method which performs learning via less but robust regions. The overall procedure of the proposed method is shown in Fig. \ref{scheme}.

\subsection{Problem Formulation}

Given a ResNet backbone as a feature extractor, and a scene recognition image $X \in \mathbb{R}^{w \times h \times 3}$ from sample set $\{ X_i, Y_i\}$, where the $Y_i$ is the label, the high-level feature map usually is obtained by:

\begin{equation}
    \mathbf{E} = ImageEncoder(X)
\end{equation}
where $\mathbf{E} \in \mathbb{R}^{c \times  d \times k}$ is a multi-channels feature map. Then, the predictive function is derived as follows:

\begin{equation}\label{cal}
    y = f(GAP(\mathbf{E}))
\end{equation}
where $GAP(\cdot)$ denotes the global average pooling operation. Our goal in scene recognition is simply to transfer to a downstream task of fine-tuning the ResNet, \textit{i.e.,} utilizing high-level semantic features for predicting the label.

\subsection{High-level Semantic Feature Selection}

We propose a feature selection method based on a mask mechanism, aiming to automatically screen high-level semantic features to improve the performance of the model. In the traditional feature extraction process, all feature maps are usually directly used for subsequent task processing. However, not all features contribute equally to the target task. In order to improve the efficiency and accuracy of the model, we introduced a mask  $\mathbf{M} \in [0,1]^{k \times d}$ based feature selection mechanism that enables the model to automatically learn how to filter out the most important high-level semantic features (1 represents preservation, 0 represents suppression). Equation \ref{cal} is derived as follows:
\begin{equation}
    y = f(GAP(\mathbf{E_s})), \mathbf{E_s} = \mathbf{E} \odot \mathbf{M},
\end{equation}
where $\odot$ represents the Hadamard product.

Through training, the model will automatically update the mask matrix $\mathbf{M}$, The parameters in $\mathbf{M}$ preserve important features while suppressing redundant or irrelevant features.

\subsection{Importance Regularization}

We expect to maximize the response of crucial high-level information with fewer regions. Then an $ L1$-norm-based importance regularization term is introduced as follows:

\begin{equation}
    \ell_{reg}=  \sum^{k \times d}_{i=1} |m_i|, m_i \in \mathbf{M}.
\end{equation}
This regularization term encourages sparsity in the mask by pushing the values of many elements $m_i$ to zero, thus, reducing the contribution of unrelated high-level semantic feature regions. The total optimization target consists of the prediction loss and the importance regularization loss, \textit{i.e.,} $\ell_{total} = \ell_{pre} + \lambda \cdot \ell_{reg}$, where $\lambda$ is a hyperparameter to adjust the importance regularization loss. The $\ell_{pre}$ is the CrossEntropy loss.

\begin{figure}[htbp]
    \centering
    \includegraphics[width=1\linewidth]{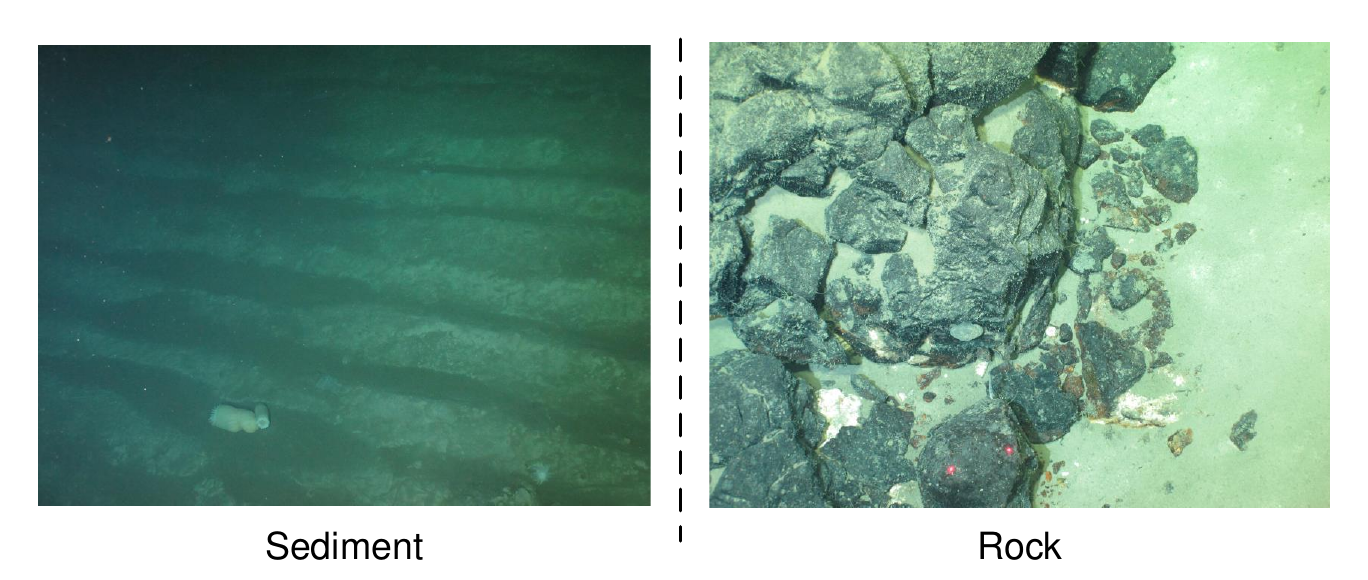}
    \caption{Example of the UGS dataset, which contains multiple common marine geology categories, e.g., sediment and rock.}
    \label{fig:enter-label}
\end{figure} 

\begin{table}[htbp]
  \vspace{-5pt}
  \renewcommand{\arraystretch}{0.9}
   \caption{Performance comparison of different models on the UGS and UCM datasets. The * behind the model indicates that it incorporates our mask matrix $\mathbf{M}$. The ``acc'' represents the average accuracy on the test set over five runs. The number following $\pm$ indicates the standard deviation of the test set accuracy across these five runs. The ``min acc'' represents the lowest accuracy on the test set observed in the five runs.
    }
  \centering
  \resizebox{1\columnwidth}{!}{
  \begin{threeparttable}
  \scriptsize 
  \renewcommand{\multirowsetup}{\centering}
  \setlength{\tabcolsep}{1.45pt}
  \begin{tabular}{c|cc|cc} 

    \toprule
    \multirow{2}{*}{Models} & \multicolumn{2}{c}{UGS} & \multicolumn{2}{c}{UCM} \\
    & acc & min acc & acc & min acc \\
    \midrule
    ResNet18        & $0.900\pm7.5e^{-4}$ & 0.900 & $0.964\pm1.5e^{-4}$ & 0.945 \\
    ResNet50        & $0.885\pm6.5e^{-4}$ & 0.850 & $0.947\pm5.4e^{-4}$ & 0.909 \\
    ResNet101       & $0.860\pm3.2e^{-3}$  & 0.750 & $0.946\pm9.0e^{-4}$ & 0.901 \\
    ResNet18*       & $\textcolor{red}{0.940}\pm1.2e^{-3}$  & 0.900 & $\textcolor{red}{0.973}\pm6.5e^{-5}$ & 0.959 \\
    ResNet50*       & $\textcolor{blue}{0.920}\pm1.4e^{-3}$  & 0.900 & $0.948\pm2.9e^{-4}$ & 0.941 \\
    ResNet101*      & $0.900\pm1.3e^{-3}$  & 0.850 & $0.951\pm1.9e^{-4}$ & 0.932 \\
    MobileNet v2    & $0.875\pm7.5e^{-4}$ & 0.850 & $0.963\pm6.6e^{-5}$ & 0.951 \\
    MobileNet v3 small & $0.680\pm1.7e^{-2}$ & 0.575 & $0.957\pm4.1e^{-4}$ & 0.917 \\
    MobileNet v3 large & $0.725\pm3.5e^{-2}$ & 0.575 & $0.956\pm4.9e^{-4}$ & 0.920 \\
    MobileNet v2*   & $0.880\pm6.0e^{-4}$ & 0.850 & $\textcolor{blue}{0.972}\pm4.7e^{-5}$ & 0.961 \\
    MobileNet v3 small* & $0.695\pm1.3e^{-2}$ & 0.575 & $0.962\pm1.0e^{-4}$ & 0.946 \\
    MobileNet v3 large* & $0.765\pm9.2e^{-3}$ & 0.575 & $0.966\pm2.3e^{-4}$ & 0.936 \\
    ViT Base 16     & $0.835\pm4.7e^{-3}$  & 0.725 & $0.930\pm9.2e^{-5}$ & 0.913 \\
    ViT Base 32     & $0.880\pm1.1e^{-3}$  & 0.875 & $0.939\pm1.9e^{-5}$ & 0.934 \\
    ViT Large 16    & $0.780\pm4.6e^{-3}$  & 0.675 & $0.886\pm6.8e^{-4}$ & 0.863 \\
    ViT Large 32    & $0.815\pm4.2e^{-3}$  & 0.725 & $0.945\pm4.4e^{-5}$ & 0.935 \\
    \bottomrule
  \end{tabular}
  \end{threeparttable}
  }
   
\label{tab:main_result}
\end{table}

\begin{figure}[htbp]
    \centering
    \includegraphics[width=1\linewidth]{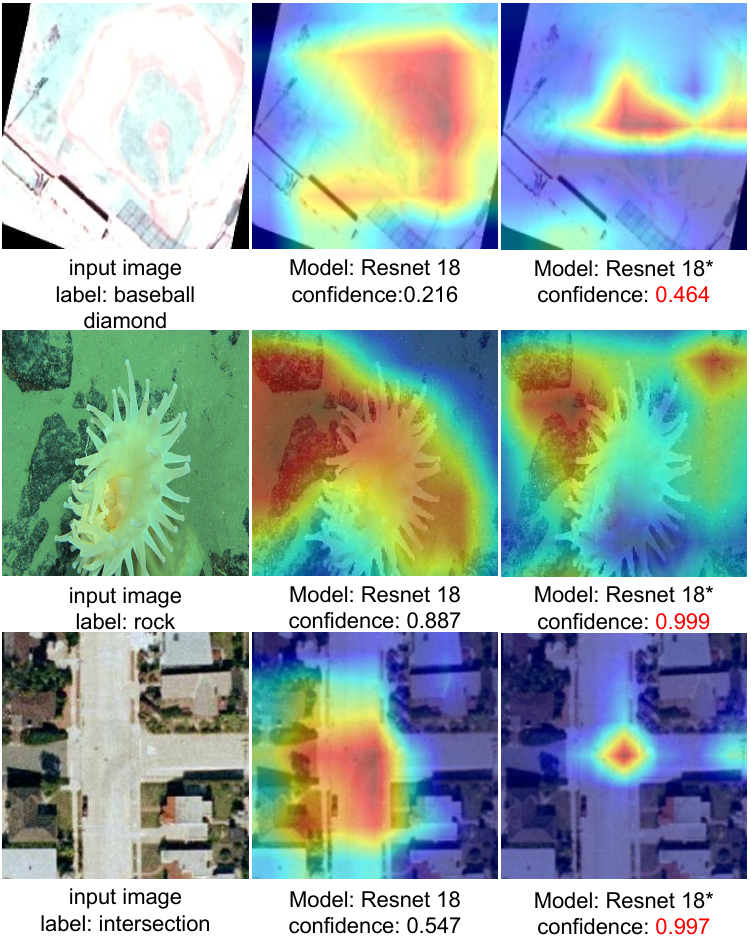}
    \caption{Visualization of changes in the attention areas of the input images before and after incorporating the mask matrix $\mathbf{M}$ in ResNet18. The * behind the model name indicates that the model incorporates our mask matrix $\mathbf{M}$.}
    \label{fig:visualize_all}
\end{figure} 

\begin{figure*}[t]
    \centering
    \subfloat[]{\includegraphics[width=0.24\textwidth]{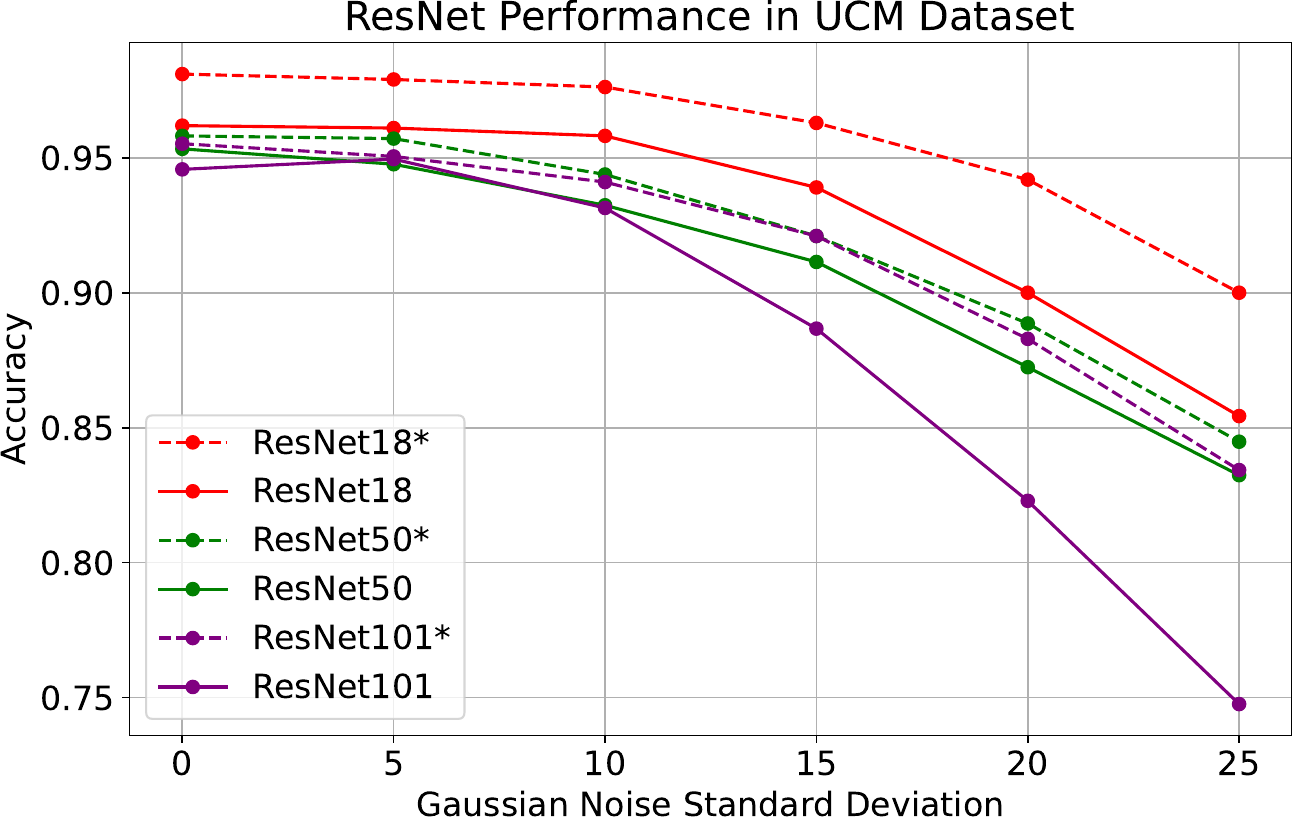}
    \label{fig:robust_1}}
    \hfil
    \subfloat[]{\includegraphics[width=0.24\textwidth]{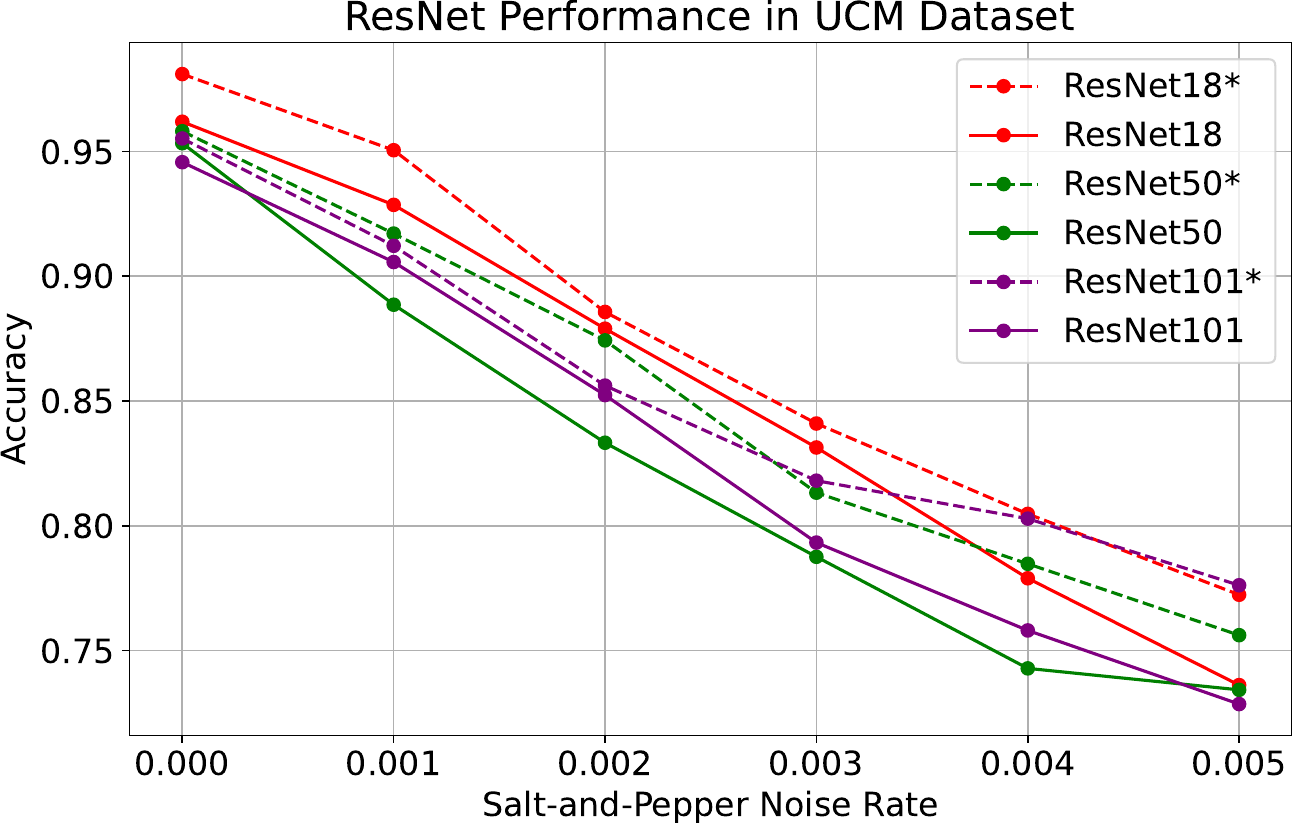}
    \label{fig:robust_2}}
    \hfil
    \subfloat[]{\includegraphics[width=0.24\textwidth]{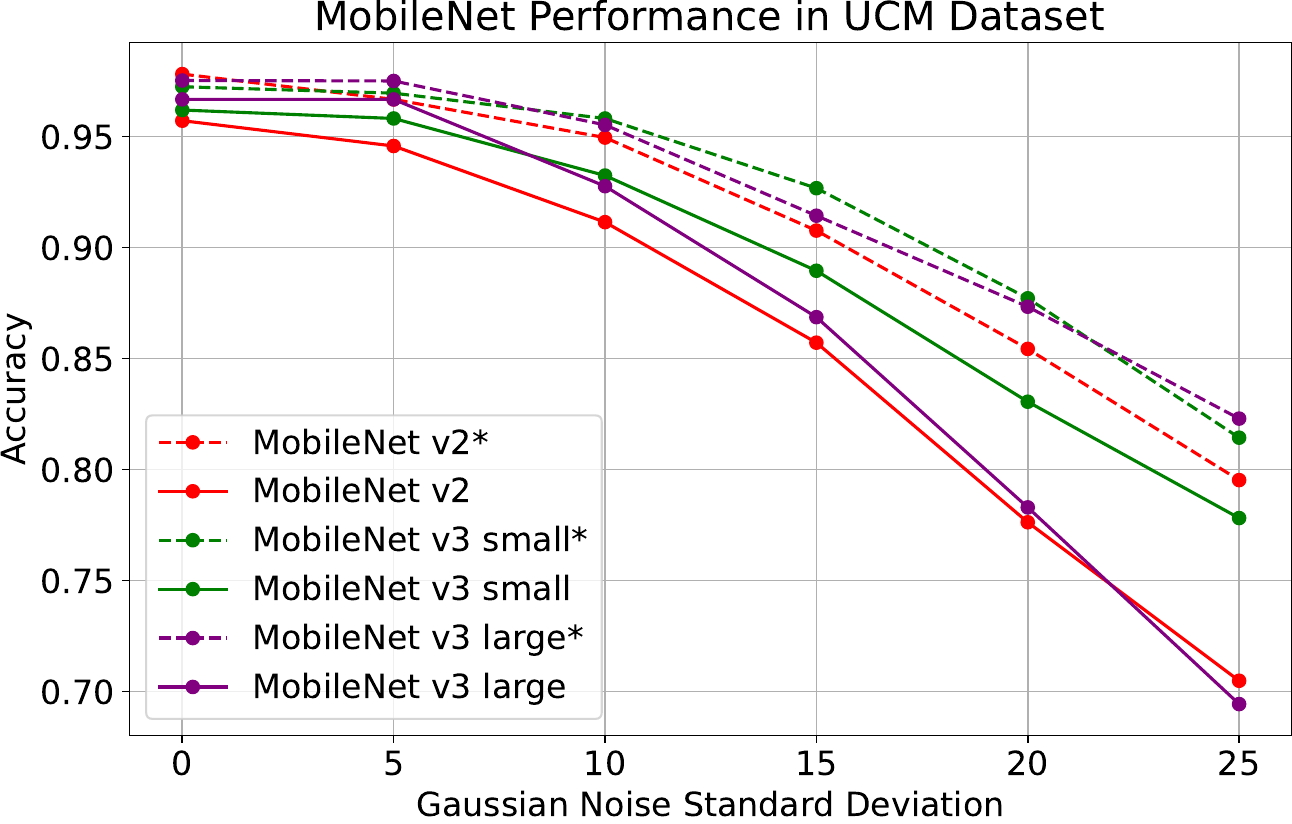}
    \label{fig:robust_3}}
    \hfil
    \subfloat[]{\includegraphics[width=0.24\textwidth]{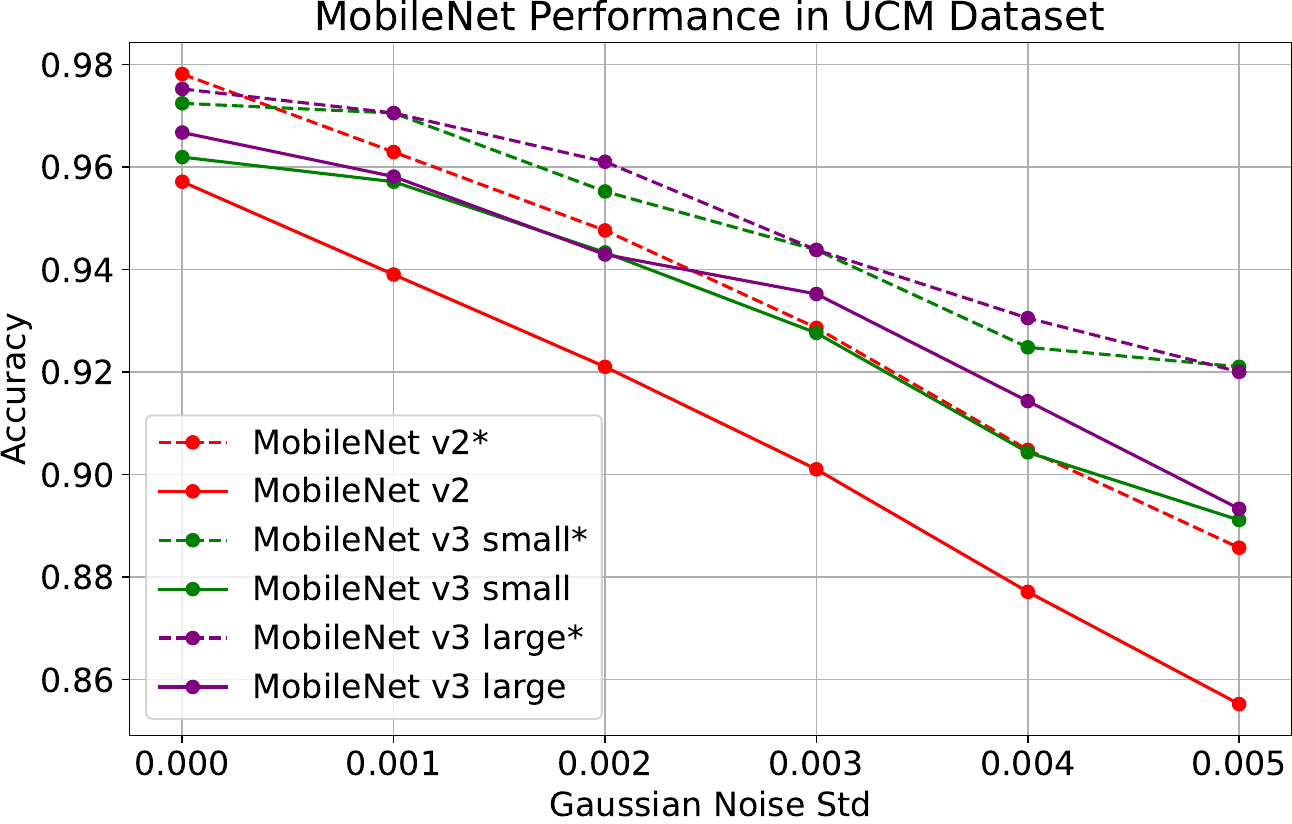}
    \label{fig:robust_4}}
    \hfil
    \caption{Comparison of model accuracy under the influence of noise. The * behind the model name indicates that the model incorporates our mask matrix $\mathbf{M}$. (a) Using ResNet-based models and Gaussian noise. (b) Using ResNet-based models and salt-and-pepper noise. (c) Using MobileNet-based models and Gaussian noise. (d) Using MobileNet-based models and salt-and-pepper noise. The experiments are conducted on the UCM dataset.}
    
    \label{fig:robust}
\end{figure*}

\begin{figure}[htbp]
    \centering
    \subfloat[]{\includegraphics[width=0.23\textwidth]{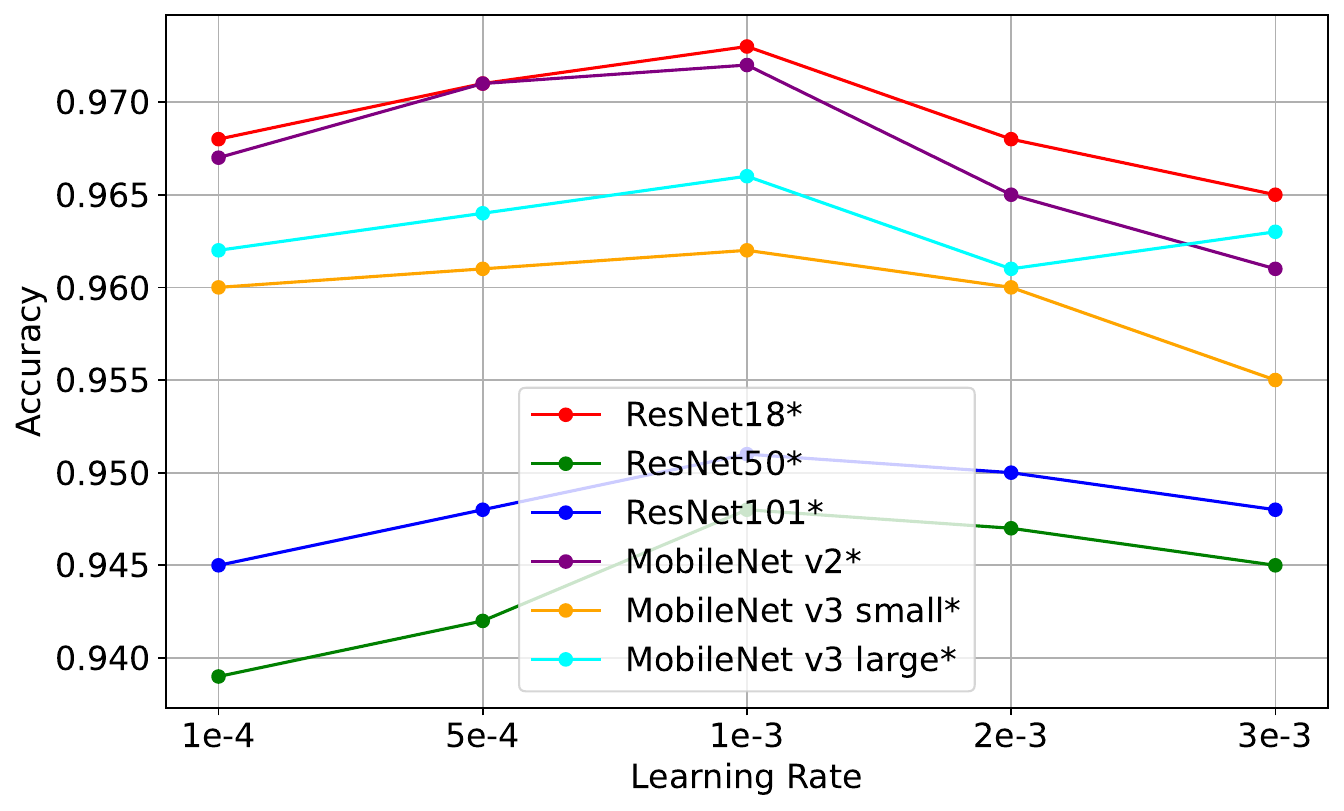}
    \label{fig:sen_1}}
    \hfil
    \subfloat[]{\includegraphics[width=0.23\textwidth]{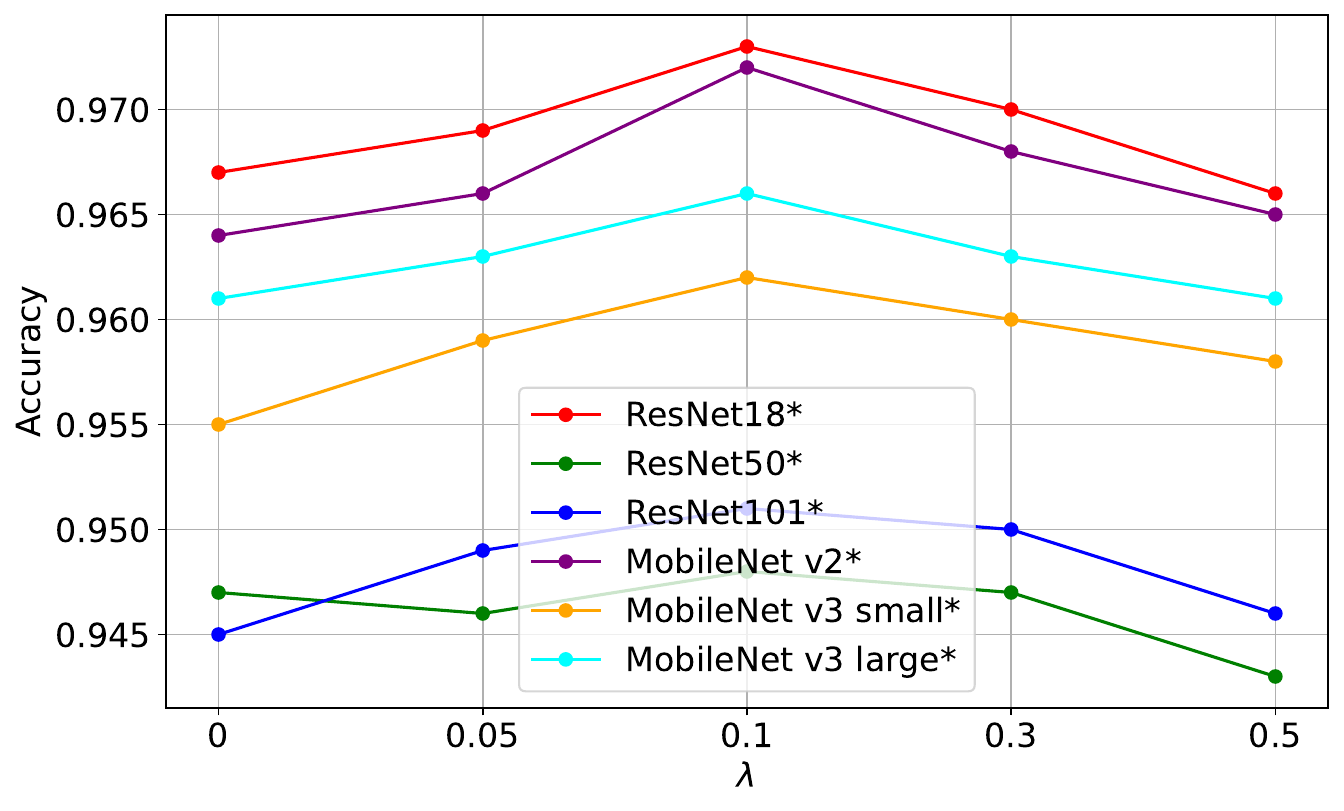}
    \label{fig:sen_2}}
    \caption{The hyperparameter sensitivity with respect to the learning rate (a) and $\lambda$ (b). The * behind the model name indicates that the model incorporates our mask matrix $\mathbf{M}$. The experiments are conducted on the UCM dataset.}
    \label{fig:sen}
\end{figure}

\section{Experiments}
This section validates the effectiveness of the proposed method through extensive experiments. First, we introduce the experimental settings. Then, we present the comparative results of multiple models and our method on two datasets. Next, we further demonstrate the effectiveness of our method through visualization experiments and robustness experiments. Finally, we analyze the hyperparameter sensitivity.

\subsection{Experiment Setting}
\paragraph{Datasets} The first dataset is our self-built UGS (Underwater Geological Scene) dataset, which contains seabed photos. These photos are taken by our Fendouzhe Striver deep-sea manned submersible using an underwater high-definition camera, with a resolution of $3648 \times 2736$. The photos are captured on the seabed of the East China Sea. The images in this dataset are divided into two categories: sediment and rock, with approximately 500 images per category, as shown in Fig. \ref{fig:enter-label}. The second dataset is the UCM dataset \cite{UCM}, which contains 21 categories of remote sensing images, with 500 images per category. Both datasets are divided into training, validation, and test sets in a ratio of 60\%, 20\%, and 20\%, respectively.

\paragraph{Baselines} We select three types of classic image classification methods for our experiments, including i) ResNet variants (ResNet18, ResNet50, ResNet101) \cite{he2016deep,cnn2021gesture}; ii) MobileNet variants (MobileNet v2, MobileNet v3 small, MobileNet v3 large) \cite{howard2017mobilenets,howard2019searching}; iii) ViT variants (ViT Base 16, ViT Base 32, ViT Large 16, ViT Large 32) \cite{alexey2020image}, where 16 and 32 represent the patch size (length and width) of ViT. 

\paragraph{Implementation Details} All models are trained on a single V100 GPU, with a batch size of 16. All images are resized to 224x224 before training and testing. $\lambda$ is set to 0.1. For all experiments, we use the Adam optimizer \cite{Adam}, with a learning rate set to 1e-4 or 1e-3. The training is conducted for up to 200 epochs, with early stopping if the validation loss does not decrease for ten consecutive epochs.

\subsection{Comparative Experimental Results}
In the comparison experiments, we use ResNet variants, MobileNet variants, VIT variants, and our proposed ResNet+mask matrix $\mathbf{M}$ method. Additionally, to validate the effectiveness of the mask matrix $\mathbf{M}$, we also incorporate it into the MobileNet variants and conduct experiments. To eliminate the influence of randomness, each model is tested five times on each dataset, and the mean accuracy, standard deviation, and minimum accuracy are recorded. The comprehensive experimental results are shown in Table \ref{tab:main_result}. The best results are highlighted in red, and the second-best results are marked in blue. As shown in Table \ref{tab:main_result}, after incorporating the mask matrix $\mathbf{M}$, the average accuracy of all models significantly improves, and the accuracy variance decreases for almost all models. This demonstrates that the proposed method enhances the predictive performance of the existing models.

\subsection{Visualization and Robustness Experiments}
\label{visual_exp}
In this section, we first show the visualization results of the models before and after incorporating the mask matrix $\mathbf{M}$. Next, we evaluate the models' robustness by testing them on the data from the test set with different types of noise added.

To qualitatively demonstrate the effect of the mask matrix $\mathbf{M}$ on the model, we use Grad-CAM \cite{Grad2017} to visualize the changes in the areas of interest for ResNet18 before and after adding the $\mathbf{M}$. As shown in Fig. \ref{fig:visualize_all}, after incorporating the $\mathbf{M}$, the model's areas of focus become more concentrated, and the confidence score of the true label is bigger.

To observe the impact of the mask matrix $\mathbf{M}$ on the model's robustness, we add Gaussian noise or salt-and-pepper noise to the test data. We then use the trained models to predict the noisy data and record the improvement in prediction accuracy after adding the mask matrix $\mathbf{M}$ to the models. For Gaussian noise, the mean is set to 0, and the variance is set to {0, 5, 10, 15, 20, 25}, with higher variance indicating stronger noise. For salt-and-pepper noise, the noise ratio is set to {0\%, 0.1\%, 0.2\%, 0.3\%, 0.4\%, 0.5\%}, representing the percentage of salt-and-pepper pixels in the total image, where a higher ratio indicates more severe noise. The experimental results are shown in Fig. \ref{fig:robust}. From the results, We can see, especially in the Gaussian noise experiments, the greater the noise, the more significant the accuracy improvement after incorporating $\mathbf{M}$.

\subsection{Hyperparameter Sensitivity}
We evaluate the hyperparameter sensitivity of our method with respect to the learning rate and $\lambda$. The results are shown in Fig. \ref{fig:sen}. We find that the model is not very sensitive to these two hyperparameters.

\section{Conclusion}

In this work, we propose a novel method that utilizes a mask-based feature selection mechanism to facilitate the model's focus on fewer yet robust high-level feature regions for the scene recognition task. An importance-based regularization is developed to further emphasize those robust areas, preventing misclassification between similar scenes. 
This method is plug-and-play and can be applied to any CNN-based method.
To simulate real-life scenarios, we conduct an underwater scene recognition dataset, called UGS, for evaluation. It contains a variety of real underwater collection scenes, covering practical situations such as biological occlusion and light blur. Extensive experimental results showcase the effectiveness and robustness of our method.

\clearpage
\newpage
\bibliography{IEEEexample}

\begin{thebibliography}{10}
\providecommand{\url}[1]{#1}
\csname url@samestyle\endcsname
\providecommand{\newblock}{\relax}
\providecommand{\bibinfo}[2]{#2}
\providecommand{\BIBentrySTDinterwordspacing}{\spaceskip=0pt\relax}
\providecommand{\BIBentryALTinterwordstretchfactor}{4}
\providecommand{\BIBentryALTinterwordspacing}{\spaceskip=\fontdimen2\font plus
\BIBentryALTinterwordstretchfactor\fontdimen3\font minus \fontdimen4\font\relax}
\providecommand{\BIBforeignlanguage}[2]{{%
\expandafter\ifx\csname l@#1\endcsname\relax
\typeout{** WARNING: IEEEtran.bst: No hyphenation pattern has been}%
\typeout{** loaded for the language `#1'. Using the pattern for}%
\typeout{** the default language instead.}%
\else
\language=\csname l@#1\endcsname
\fi
#2}}
\providecommand{\BIBdecl}{\relax}
\BIBdecl

\bibitem{Zhao2018}
T.~Zhang and X.~Huang, ``Monitoring of urban impervious surfaces using time series of high-resolution remote sensing images in rapidly urbanized areas: A case study of shenzhen,'' \emph{IEEE Journal of Selected Topics in Applied Earth Observations and Remote Sensing}, vol.~11, no.~8, pp. 2692--2708, 2018.

\bibitem{2022Bh}
R.~Bhuvaneswari, T.~Surya, T.~Srikanth, and R.~Balaji, ``A novel approach for underwater object detection through deep intense-net for ocean conservation systems,'' in \emph{OCEANS 2022 - Chennai}, 2022, pp. 1--9.

\bibitem{2021GU}
G.~Gubnitsky and R.~Diamant, ``A multispectral target detection in sonar imagery,'' in \emph{OCEANS 2021: San Diego – Porto}, 2021, pp. 1--5.

\bibitem{2022Qi}
H.~Qi and Y.~Tang, ``Autonomous underwater rescue technology,'' in \emph{OCEANS 2022 - Chennai}, 2022, pp. 1--4.

\bibitem{aid}
G.-S. Xia, J.~Hu, F.~Hu, B.~Shi, X.~Bai, Y.~Zhong, L.~Zhang, and X.~Lu, ``Aid: A benchmark data set for performance evaluation of aerial scene classification,'' \emph{IEEE Transactions on Geoscience and Remote Sensing}, vol.~55, no.~7, pp. 3965--3981, 2017.

\bibitem{Eurosat}
P.~Helber, B.~Bischke, A.~Dengel, and D.~Borth, ``Eurosat: A novel dataset and deep learning benchmark for land use and land cover classification,'' \emph{IEEE Journal of Selected Topics in Applied Earth Observations and Remote Sensing}, vol.~12, no.~7, pp. 2217--2226, 2019.

\bibitem{UCM}
Y.~Yang and S.~Newsam, ``Bag-of-visual-words and spatial extensions for land-use classification,'' 11 2010, pp. 270--279.

\bibitem{Liu2016}
\BIBentryALTinterwordspacing
Z.~Liu, Y.~Zhang, X.~Yu, and C.~Yuan, ``Unmanned surface vehicles: An overview of developments and challenges,'' \emph{Annu. Rev. Control.}, vol.~41, pp. 71--93, 2016. [Online]. Available: \url{https://api.semanticscholar.org/CorpusID:36594615}
\BIBentrySTDinterwordspacing

\bibitem{Yuh2011}
J.~Yuh, G.~Marani, and D.~R. Blidberg, ``Applications of marine robotic vehicles,'' \emph{Intelligent Service Robotics}, vol.~4, pp. 221--231, 2011.

\bibitem{underwater2022utd}
J.~Wang and N.~Yu, ``Utd-yolov5: a real-time underwater targets detection method based on attention improved yolov5,'' \emph{arXiv preprint arXiv:2207.00837}, 2022.

\bibitem{Chaib2017}
S.~Chaib, H.~Liu, Y.~Gu, and H.~Yao, ``Deep feature fusion for vhr remote sensing scene classification,'' \emph{IEEE Transactions on Geoscience and Remote Sensing}, vol.~55, no.~8, pp. 4775--4784, 2017.

\bibitem{wang2024towards}
J.~Wang, W.~Qiang, X.~Su, C.~Zheng, F.~Sun, and H.~Xiong, ``Towards task sampler learning for meta-learning,'' \emph{International Journal of Computer Vision}, pp. 1--31, 2024.

\bibitem{Grad2017}
R.~R. Selvaraju, M.~Cogswell, A.~Das, R.~Vedantam, D.~Parikh, and D.~Batra, ``Grad-cam: Visual explanations from deep networks via gradient-based localization,'' in \emph{2017 IEEE International Conference on Computer Vision (ICCV)}, 2017, pp. 618--626.

\bibitem{2019H}
R.~Minetto, M.~Pamplona~Segundo, and S.~Sarkar, ``Hydra: An ensemble of convolutional neural networks for geospatial land classification,'' \emph{IEEE Transactions on Geoscience and Remote Sensing}, vol.~57, no.~9, pp. 6530--6541, 2019.

\bibitem{2019ww}
Q.~Wang, S.~Liu, J.~Chanussot, and X.~Li, ``Scene classification with recurrent attention of vhr remote sensing images,'' \emph{IEEE Transactions on Geoscience and Remote Sensing}, vol.~57, no.~2, pp. 1155--1167, 2019.

\bibitem{2015cheng}
G.~Cheng, J.~Han, L.~Guo, and T.~Liu, ``Learning coarse-to-fine sparselets for efficient object detection and scene classification,'' in \emph{2015 IEEE Conference on Computer Vision and Pattern Recognition (CVPR)}, 2015, pp. 1173--1181.

\bibitem{wang2022ssd}
J.~Wang and N.~Yu, ``Ssd-faster net: A hybrid network for industrial defect inspection,'' \emph{arXiv preprint arXiv:2207.00589}, 2022.

\bibitem{2016Deep}
K.~He, X.~Zhang, S.~Ren, and J.~Sun, ``Deep residual learning for image recognition,'' \emph{IEEE}, 2016.

\bibitem{2009L}
J.~Deng, W.~Dong, R.~Socher, L.-J. Li, K.~Li, and L.~Fei-Fei, ``Imagenet: A large-scale hierarchical image database,'' in \emph{2009 IEEE Conference on Computer Vision and Pattern Recognition}, 2009, pp. 248--255.

\bibitem{Liu2024SCECNetSF}
\BIBentryALTinterwordspacing
X.~Liu, W.~Wu, Z.~Hu, and Y.~Sun, ``Scecnet: self-correction feature enhancement fusion network for remote sensing scene classification,'' \emph{Earth Science Informatics}, 2024. [Online]. Available: \url{https://api.semanticscholar.org/CorpusID:271183132}
\BIBentrySTDinterwordspacing

\bibitem{s2023}
C.~Tang, X.~Zheng, and C.~Tang, ``Adaptive discriminative regions learning network for remote sensing scene classification,'' \emph{Sensors}, vol.~23, no.~2, 2023.

\bibitem{2017MMMM}
A.~Mahmood, M.~Bennamoun, S.~An, and F.~Sohel, ``Resfeats: Residual network based features for image classification,'' in \emph{2017 IEEE International Conference on Image Processing (ICIP)}, 2017, pp. 1597--1601.

\bibitem{Phapale2021}
A.~Phapale, P.~Kasture, K.~Katkar, O.~V. Karale, and A.~Deshmukh, ``Underwater image enhancement with a deep residual framework,'' \emph{International Journal of Scientific Research in Science and Technology}, 2021.

\bibitem{he2016deep}
K.~He, X.~Zhang, S.~Ren, and J.~Sun, ``Deep residual learning for image recognition,'' in \emph{Proceedings of the IEEE conference on computer vision and pattern recognition}, 2016, pp. 770--778.

\bibitem{cnn2021gesture}
W.~Jingyao, Y.~Naigong, and E.~Firdaous, ``Gesture recognition matching based on dynamic skeleton,'' in \emph{2021 33rd Chinese Control and Decision Conference (CCDC)}.\hskip 1em plus 0.5em minus 0.4em\relax IEEE, 2021, pp. 1680--1685.

\bibitem{howard2017mobilenets}
A.~G. Howard, ``Mobilenets: Efficient convolutional neural networks for mobile vision applications,'' \emph{arXiv preprint arXiv:1704.04861}, 2017.

\bibitem{howard2019searching}
A.~Howard, M.~Sandler, G.~Chu, L.-C. Chen, B.~Chen, M.~Tan, W.~Wang, Y.~Zhu, R.~Pang, V.~Vasudevan \emph{et~al.}, ``Searching for mobilenetv3,'' in \emph{Proceedings of the IEEE/CVF international conference on computer vision}, 2019, pp. 1314--1324.

\bibitem{alexey2020image}
D.~Alexey, ``An image is worth 16x16 words: Transformers for image recognition at scale,'' \emph{arXiv preprint arXiv: 2010.11929}, 2020.

\bibitem{Adam}
D.~P. Kingma and J.~Ba, ``Adam: {A} method for stochastic optimization,'' \emph{ICLR}, 2015.

\end{thebibliography}
\bibliographystyle{IEEEtran}

\end{document}